%% file: main.tex
\documentclass[10pt,twocolumn,letterpaper]{article}

\usepackage{wacv}              


\definecolor{wacvblue}{rgb}{0.21,0.49,0.74}
\usepackage[pagebackref,breaklinks,colorlinks,allcolors=wacvblue]{hyperref}
\usepackage{subcaption}

\def\wacvPaperID{1586} 
\def\confName{WACV}
\def\confYear{2026}

\title{StreetView-Waste: A Multi-Task Dataset for Urban Waste Management}

\author{
Diogo J. Paulo$^{1,2}$, João Martins$^{1}$, Hugo Proença$^{1,2}$, João C. Neves$^{1,3}$\\[4pt]
\small $^{1}$University of Beira Interior, Portugal \quad \small $^{2}$IT: Instituto de Telecomunicações \small \quad $^{3}$NOVA LINCS\\
\small \texttt{diogo.paulo@ubi.pt}
}

\usepackage[table]{xcolor}
\usepackage{multirow}
\usepackage{cuted}
\usepackage{graphicx}  
\begin{document}
\maketitle
\input{0_abstract}    
\input{1_intro}
\input{2_related_work}
\input{3_dataset}
\input{4_proposed_strategies}
\input{5_experiments}
\input{6_conclusions}

{
    \small
    \bibliographystyle{ieeenat_fullname}
    \bibliography{main}
}

\end{document}

%% file: 0_abstract.tex

\begin{strip}
    \centering
    \includegraphics[width=\linewidth]{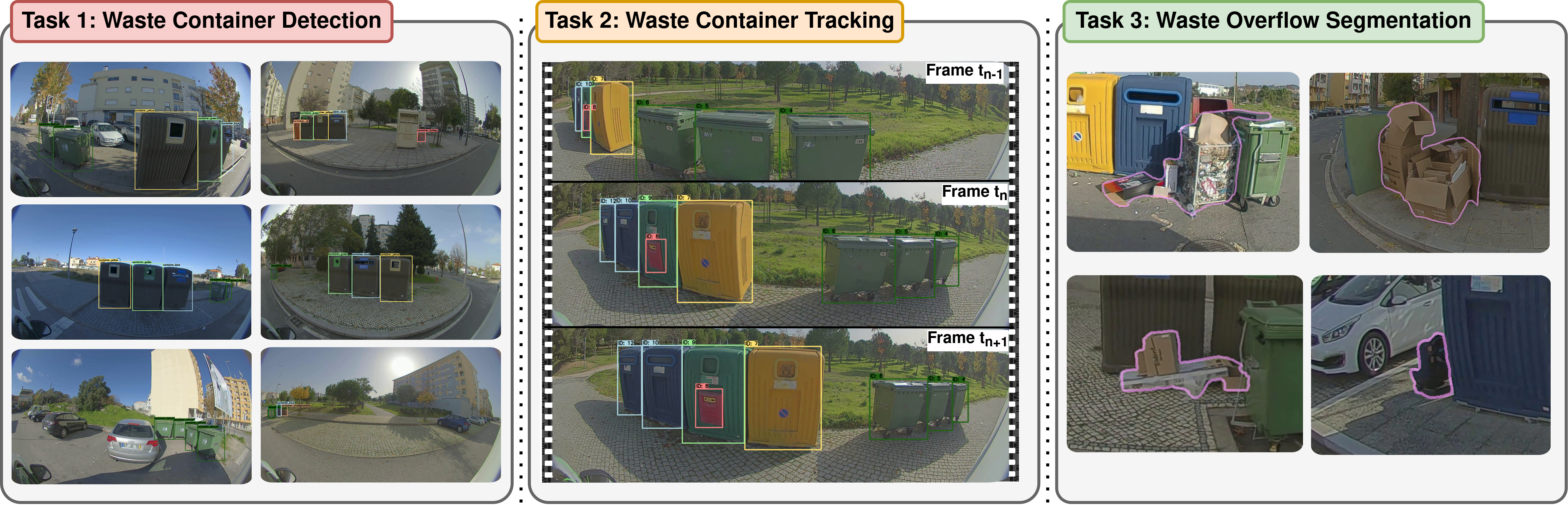}
    \captionof{figure}{We introduce \textbf{StreetView-Waste}, the first fisheye image dataset tailored for urban waste analysis. Captured using two 180$^{\circ}$ field of view cameras, the dataset mirrors the settings of real urban waste collection, providing high-quality annotations for three core tasks: 2D object detection, object tracking, and instance segmentation. These tasks are critical for logistics, with detection and overflow segmentation enabling status assessment, while tracking is essential for mapping municipal assets and optimizing collection routes. StreetView-Waste serves as a foundation for developing robust, real-world waste analysis models.}
    \label{fig:teaser}
\end{strip}
\begin{abstract}
Urban waste management remains a critical challenge for the development of smart cities. Despite the growing number of litter detection datasets, the problem of monitoring overflowing waste containers — particularly from images captured by garbage trucks — has received little attention. While existing datasets are valuable, they often lack annotations for specific container tracking or are captured in static, decontextualized environments, limiting their utility for real-world logistics. To address this gap, we present StreetView-Waste, a comprehensive dataset of urban scenes featuring litter and waste containers. The dataset supports three key evaluation tasks: (1) waste container detection, (2) waste container tracking, and (3) waste overflow segmentation.
Alongside the dataset, we provide baselines for each task by benchmarking state-of-the-art models in object detection, tracking, and segmentation. Additionally, we enhance baseline performance by proposing two complementary strategies: a heuristic-based method for improved waste container tracking and a model-agnostic framework that leverages geometric priors to refine litter segmentation.
Our experimental results show that while fine-tuned object detectors achieve reasonable performance in detecting waste containers, baseline tracking methods struggle to accurately estimate their number; however, our proposed heuristics reduce the mean absolute counting error by 79.6\%. Similarly, while segmenting amorphous litter is challenging, our geometry-aware strategy improves segmentation mAP@0.5 by 27\% on lightweight models, demonstrating the value of multimodal inputs for this task. Ultimately, StreetView-Waste provides a challenging benchmark to encourage research into real-world perception systems for urban waste management.
\end{abstract}

%% file: 1_intro.tex
\vspace{-1em}
\section{Introduction}
\label{sec:intro}
Effective urban waste management is critical for public health and environmental sustainability. While waste collection has been done manually at fixed-schedule time slots, there is a growing interest in leveraging computer vision for either inferring statistics of waste distribution or even automating waste collection. This explains the growing number of computer vision datasets for litter detection. However, existing datasets often focus on general litter classification or lack the specific, dynamic context needed for real-world logistics. To bridge this gap, this paper introduces StreetView-Waste, a comprehensive, publicly available dataset comprising 36,478 fisheye images collected in real-world urban settings. The data was captured from a collection vehicle over several weeks to encompass a wide range of real-world urban settings, weather patterns, and lighting conditions, ensuring high environmental diversity. Our dataset is uniquely designed to support three key computer vision tasks: (1) \textbf{waste container detection}, annotated with over 71,000 bounding boxes across diverse container type, (2) \textbf{waste container tracking}, which includes nearly 400 unique container tracks to evaluate temporal consistency, and (3) \textbf{waste overflow segmentation}, featuring over 5,000 detailed instance masks of litter surrounding containers.

Furthermore, to demonstrate the value of our dataset for exploring novel research directions, we focus on the most challenging tasks: waste container tracking and overflowing waste segmentation. We propose and evaluate a model-agnostic approach that enhances segmentation models by fusing RGB data with geometric information and a heuristic-based method for improving waste container tracking. While traditional approaches~\cite{balmik2023vision43,de2020yolo44, kulkarni2019waste41, ouguz2023determining42} rely solely on RGB input information or from IoT sensors, we enhance the power of segmentation models by introducing geometric information and developing a model-agnostic solution. We conduct extensive experiments on a variety of state-of-the-art models, including detection, tracking, and segmentation methods. Our results demonstrate significant performance gains in segmentation accuracy when combined with geometric information. Nonetheless, the accuracy of our system is inherently connected to the quality of depth and normal predictions, which are susceptible to noise; hence, we also discuss the limitations exposed by our diagnostic tools. Our findings confirm that tracking waste containers and segmenting overflowing litter in street-level scenes remains an open problem. To this end, we present StreetView-Waste as a challenging new benchmark, paving the way to accelerate research in this domain. While our proposed strategies for segmentation and tracking achieve significant gains over baselines — reducing container counting error by 79.6\% and boosting overflow segmentation mAP@0.5 by 27\% on lightweight models — the results show that there is still room for improvement. We believe our dataset and initial findings will serve as a valuable resource for the community to develop and validate the next generation of solutions for automated urban waste management.
Our main contributions are as follows:

\begin{itemize}

    \item \textbf{A Novel, Large-Scale Dataset for Urban Waste Management.} We introduce and make publicly available StreetView-Waste, a new, large-scale, multi-task dataset of over 36,000 fisheye images captured from the street-level perspective of waste collection vehicles.

    \item \textbf{A Comprehensive Benchmark on Litter and Waste Container Analysis.} We provide an extensive benchmark of state-of-the-art models on the proposed dataset for three key tasks: waste container detection, tracking and counting, and waste overflow segmentation. This work establishes a thorough performance benchmark, providing a reference for future research in this domain.
    
    \item \textbf{Innovative Strategies for Segmentation and Tracking.} We propose and validate two distinct strategies that enhance general-purpose models for critical waste management tasks. The first is a model-agnostic method that improves overflow segmentation by fusing RGB data with geometric information. The second is a novel heuristic-based approach that increases the accuracy and robustness of waste container tracking and counting.
\end{itemize}
The full dataset and code will be made publicly available.

%% file: 2_related_work.tex
\section{Related Work}
\label{sec:related_work}
\noindent\textbf{Litter Detection Datasets.}
Several works have focused attention on the detection of litter. Initial studies focused on waste classification, aiming to sort materials for recycling~\cite{trashnet1, GLASSENSEVISION2, wasteclassificationdata3, wasteimagesushi4, openlittermap5, spotgarbage6, realwaste7, PortlandStateSinghRECYCLE8, tidy9, garbageclassification10, cigarette_butt_dataset11, deepseawaste12, wadaba13, wadaba14}. While these are vital for sorting applications, these datasets typically feature decontextualized images of isolated objects with simple backgrounds, limiting their utility for in-the-wild scene understanding. To address real-world scenarios, research has progressed towards litter detection and segmentation in complex environments. Among these works,~\cite{taco15, uavaste16, bepli17, mjuwaste18, bashkirova2022zerowaste, lots22} provide instance segmentation masks for different litter categories across various settings. Many other datasets provide bounding box annotations for litter detection, including~\cite{trashicra19,wadaba13,wadaba14,trashbox20}. For underwater environments,~\cite{trashcan21,deepseawaste12} offer benchmarks for detecting marine debris. For aerial surveillance, UAVVaste~\cite{uavaste16} provides data captured from drones. Others address unique surface conditions, such as LOTS~\cite{lots22} for segmenting litter on sandy beaches, or specific object types, like the BePLi Dataset v1~\cite{bepli17} for plastic litter. Although these domain-specific datasets are invaluable, they do not address the distinct perceptual challenges of ground-level urban waste management; a critical gap remains from the operational perspective of municipal service vehicles. Furthermore, large-scale datasets like WoodScape~\cite{woodscape23} use fisheye cameras, commonly found on such vehicles, but are annotated for autonomous driving tasks (e.g., vehicles, pedestrians), and entirely lack the labels required for waste management. The task of monitoring waste containers for efficient collection logistics remains underexplored. While some datasets include waste containers as a class~\cite{detectingoverfilledbins24, wastecontainers25, valente2019computer26}, no existing resource provides a comprehensive benchmark for the joint tasks of waste container detection, multi-object tracking from a moving vehicle, and waste overflow segmentation. Our work is explicitly designed to fill this void. To the best of our knowledge, StreetView-Waste is the first large-scale dataset to combine a street-view fisheye perspective with rich, multi-task annotations for container detection, tracking, and overflow segmentation, thereby fostering the development of smart waste management systems. For a direct comparison, we summarize the key statistics of these datasets, including image and class counts, as well as the type of task, in a comparative table in the supplementary material (Table 1).

\noindent\textbf{Automated Litter Analysis.}
The datasets previously mentioned have enabled the benchmarking of various models for automated litter analysis, focusing on object detection and segmentation. For waste and container detection, several models have been employed as benchmarks~\cite{zong2023detrs33, carion2020end34, tan2020efficientdet35, girshick2015fast36, ren2015faster37, li2019scale38, yolo, liu2016ssd}. Two-stage models like Faster R-CNN~\cite{ren2015faster37} provide high accuracy and are often used for baseline evaluations on datasets like Trash-ICRA19~\cite{trashicra19} and TrashCan~\cite{trashcan21}. For applications requiring real-time performance, such as on-vehicle processing, one-stage detectors like the YOLO family~\cite{yolo} and SSD~\cite{liu2016ssd} are predominantly used due to their computational efficiency. Regarding object segmentation tasks, instance segmentation models have been adopted to produce fine-grained, pixel-level masks. The typical approach for high-quality segmentation is Mask R-CNN~\cite{he2017mask27}, which has been benchmarked on datasets like TACO~\cite{taco15} and TrashCan~\cite{trashcan21}. For real-time scenarios, faster models such as YOLACT~\cite{bolya2019yolact28}, SOLO~\cite{wang2020solo30}, and its successor SOLOv2~\cite{wang2020solov2_31} have been developed. More recently, vision transformers have set a new standard, with models like Mask2Former~\cite{cheng2022masked32} unifying panoptic, instance, and semantic segmentation into a single powerful framework.

%% file: 3_dataset.tex
\section{Dataset}
\label{sec:dataset}
StreetView-Waste comprises images of waste containers captured under various real-world conditions, as well as instances of overflowing waste and litter in the surrounding areas.  Our dataset's primary value lies in its challenging scenarios, such as severe occlusions by passing vehicles, illumination changes, and the wide range of field of view, which often cause state-of-the-art models to fail. To situate our contribution within the existing landscape, we provide a detailed comparison with other relevant public datasets in the supplementary material (Table 1) and a figure illustrating several images of StreetView-Waste (Figure 1).

\subsection{Dataset Acquisition}
To ensure a high degree of environmental and temporal diversity, our dataset was collected across multiple, distinct sessions on different days. These recording sessions were intentionally varied, taking place in both the morning and afternoon under different weather conditions, ranging from bright, direct sunlight creating harsh shadows to overcast skies with diffuse, low-contrast lighting. This strategic collection ensured that StreetView-Waste captures a wide spectrum of real-world operational scenarios, avoiding biases that might arise from a single, short-term collection period. Our dataset was captured from a vehicle equipped with two fisheye cameras, which were mounted on the vehicle's flanks, each providing a 180$^{\circ}$ field of view to ensure wide environmental coverage. Raw video sequences were recorded at 1920$\times$1080 resolution and 30 fps. To minimize temporal correlation between frames and ensure a diverse training and evaluation set, we applied a uniform subsampling procedure, resulting in an effective frame rate of 10 fps for annotation. To maximize reproducibility, the full set of camera intrinsics and distortion coefficients is provided in the supplementary material (Table 2), estimated using the generic fisheye model of Kannala and Brandt~\cite{kannala2006generic}. No rectification or cropping was applied, as the raw fisheye projection preserves critical context from both road and sidewalk regions.

\subsection{Tasks and Metrics}
Our StreetView-Waste dataset is designed to benchmark three critical tasks for automated waste management. We establish baselines for each using state-of-the-art models and standard evaluation protocols.

\noindent\textbf{Waste Container Detection.} The first task evaluates the ability of models to perform multi-class detection of seven distinct container types from individual frames. Accurate detection is the foundational step for any subsequent analysis, such as tracking or status assessment. To establish robust baselines, we evaluate two distinct state-of-the-art paradigms. We use a high-performance, frame-based detector, YOLOv11~\cite{khanam2024yolov11_29}, representing the efficiency required for real-time, on-vehicle applications, and a video object detector, DiffusionVID~\cite{roh2023diffusionvid46}, which uses temporal information across frames to potentially improve detection robustness and consistency.
Performance is measured using the standard mean average precision (mAP) metric.

\noindent\textbf{Multi-Object Tracking and Counting.}
This task addresses the challenge of associating container detections across consecutive frames to maintain unique identities, which is an important task for route optimization and mapping municipal assets. We benchmark two strong tracking-by-detection baselines: ByteTrack~\cite{zhang2022bytetrack47} and BoT-SORT~\cite{botsort}. Their ability to use low-confidence detections to handle occlusions is particularly relevant for cluttered street-view scenes where containers are frequently occluded by vehicles. Tracking performance is assessed by analyzing two aspects: the tracking accuracy and object counting estimation. In the former, we adopt a) the standard multiple object tracking accuracy (MOTA) proposed in~\cite{bernardin2008evaluatingmota}; b) the identity f1 score (IDF1), introduced in~\cite{ristani2016performanceidf1}, which is a metric that is more sensitive to long-term identity preservation compared to MOTA; and c) the higher order tracking accuracy (HOTA)~\cite{luiten2021hota}. For counting accuracy, we use the mean absolute error (MAE), root mean square error (RMSE) between the predicted count and the ground truth, mean absolute percentage error (MAPE) for providing a normalized estimation of counting deviation, and the sum of absolute count differences per sequence (SAD).

\noindent\textbf{Overflowing Waste Segmentation.}
This task aims to produce pixel-accurate instance masks for unstructured waste spilling out of containers. Although operational alarms could be binary, pixel masks are required to estimate overflow volume and prioritize routes, which is central to the logistics goal; hence, we keep overflowing waste segmentation as the primary benchmark. Precise segmentation is a prerequisite for quantifying overflow volume and triggering collection alerts. To achieve this, we benchmark a diverse set of architectures, offering a comprehensive overview of how various architectural biases address this task. We evaluate the segmentation quality using the mean average precision (mAP) over masks and the boundary intersection over union (B-IoU).

\subsection{Statistics per Task}
To facilitate rigorous and reproducible research, we provide standardized training, validation, and test splits for each task. These splits are performed at the video level to prevent data leakage, ensuring that frames from the same recording do not appear in different sets. The full StreetView-Waste dataset comprises a total of 36,478 images, captured to maximize environmental diversity. Of these, 14,219 images contain labeled objects, forming the core of our benchmarks. The statistics for each specific task are detailed below. Our dataset covers seven municipal container types reflecting common European curbside collection. Default containers are general-purpose bins deployed on most streets. Green containers are for glass. Blue containers collect paper and cardboard. Yellow containers handle lightweight packaging (plastics and metals). Biodegradable containers are for organic waste. Oil containers are dedicated units for household cooking oil due to environmental risk. Battery containers are compact drop-boxes for small batteries, often narrow and frequently occluded by street furniture. This variety mirrors real segregation rules and yields strong class imbalance for minority streams, which we capture in StreetView-Waste. An additional per-class analysis of the size and location of the containers in the image is provided in the supplementary material (Figure 2).
\begin{figure}[ht]
    \centering
    \begin{subfigure}[b]{0.48\linewidth}
        \centering
        \includegraphics[width=\linewidth]{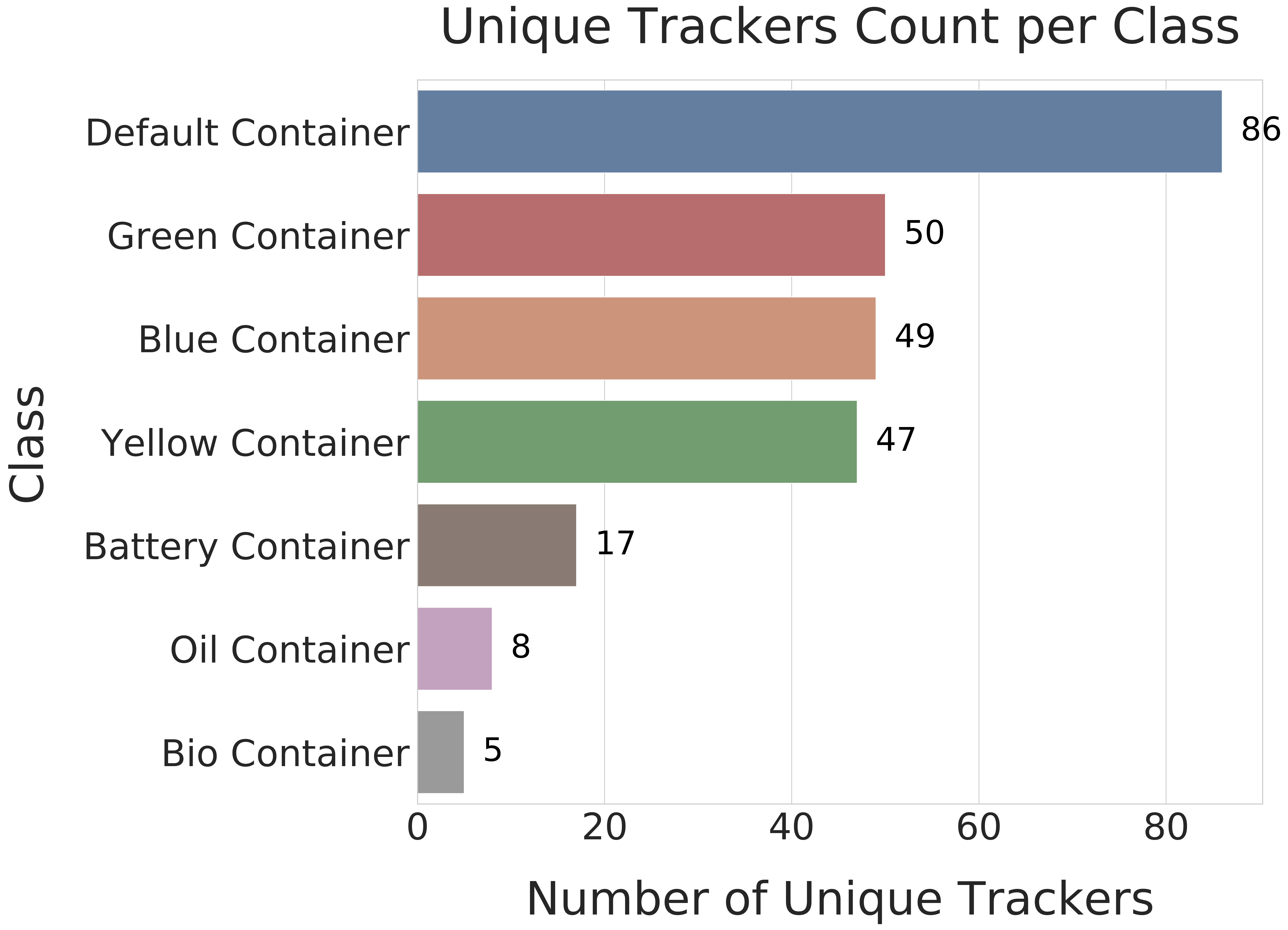}
        \caption{Unique tracks per class.}
        \label{fig:stats_tracking}
    \end{subfigure}
    \hfill
    \begin{subfigure}[b]{0.48\linewidth}
        \centering
        \includegraphics[width=\linewidth]{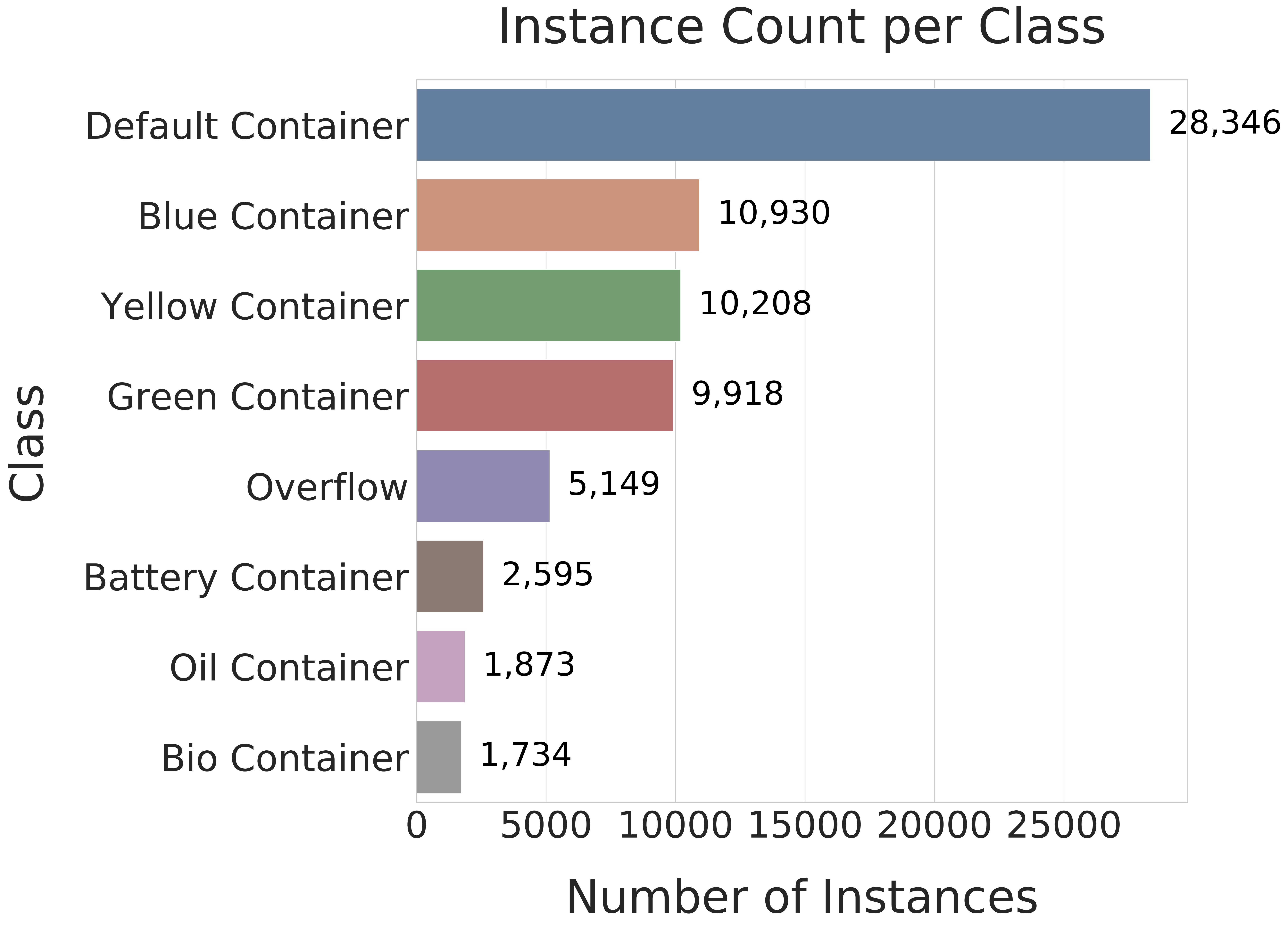}
        \caption{Total instances per class.}
        \label{fig:stats_detection}
    \end{subfigure}

    \caption{\textbf{Class distribution statistics for our StreetView-Waste dataset.} (a) Distribution of the 376 unique container tracks, highlighting the long-tail nature of the tracking task. (b) Distribution of the 71,170 total annotated instances for the detection task.}
    \label{fig:dataset_class_distributions}
\end{figure}
For waste container detection, the dataset is substantial, featuring 71,170 annotated instances across seven classes, as shown in Figure \ref{fig:stats_detection}. This scale provides a robust foundation for developing and evaluating object detectors in complex urban environments, capturing a wide diversity of container appearances and scenarios. The multi-object tracking and counting benchmark is built upon dense annotations identifying 376 unique container identities. As illustrated in Figure \ref{fig:stats_tracking}, these tracks exhibit a realistic long-tail class distribution. While common types, such as the default container, are abundant, minority classes, such as the biodegradable container and oil container, present a challenging real-world scenario for evaluating long-term tracking and re-identification. Finally, the overflowing waste segmentation task is supported by a dedicated set of annotations focusing on unstructured waste. It includes 5,149 fine-grained, pixel-level instance masks for litter and overflow. These annotations are distributed across 4,197 positive images (with the presence of waste). The complete segmentation benchmark contains 7,230 images, including negative samples, to ensure a robust evaluation of a model's ability to identify and segment overflowing waste precisely.


\subsection{Privacy Considerations and Data Integrity}
Our dataset, captured on public streets, contains faces and license plates, which are subject to the GDPR. While we considered anonymization methods like GAN-based replacement~\cite{GDPR2}, we concluded that any alteration poses a significant risk to the data's scientific utility, as artifacts could be misidentified as litter or occlude key features. Therefore, to avoid introducing harmful domain shifts, we decide to preserve the original data, a position supported by the creators of other large-scale datasets~\cite{GDPR1, woodscape23}. Consequently, access will be managed through a formal data license agreement that restricts use to academic research and requires users to comply with GDPR. This approach balances privacy obligations with the need for high-quality, unaltered data. Moreover, the data collection was conducted exclusively in public spaces where no additional ethics board approval was required under national regulations. Nevertheless, the study protocol, including privacy handling, was reviewed internally to ensure GDPR compliance. For illustrative purposes in this paper, all exemplar figures have been redacted to prevent identification.

%% file: 4_proposed_strategies.tex
\section{Proposed Strategies}
\label{sec:proposed_methods}
Our initial benchmarking experiments, detailed in Section \ref{sec:experiments}, reveal varying levels of difficulty across the three evaluation tasks. We observe that fine-tuned object detectors achieve reasonable performance on waste container detection. However, the more complex tasks of tracking and segmentation expose significant limitations in current state-of-the-art models when applied to our challenging street-view domain. For tracking, standard methods struggle with frequent occlusions and the visual similarity of containers, resulting in high rates of identity switches and track fragmentation. For segmentation, delineating amorphous, overflowing waste from its container and the cluttered urban background proves exceptionally challenging for models that rely solely on RGB data. To address these specific challenges, we introduce two complementary, model-agnostic strategies designed to enhance the performance of off-the-shelf models. These strategies target the distinct failure modes observed in tracking and segmentation, respectively.

\subsection{Heuristic-based Tracking Refinement}
While tracking-by-detection frameworks like ByteTrack~\cite{zhang2022bytetrack47} or BoT-SORT~\cite{botsort} provide a strong foundation, their performance degrades in scenarios with prolonged occlusions or when containers briefly exit and re-enter the camera's field of view. To mitigate these errors, we introduce a set of heuristics not only as a practical method to improve tracking and counting accuracy, but also as a diagnostic tool that benchmarks the specific failure modes of existing trackers in this challenging vehicular context. These rules are designed to enforce domain-specific constraints based on the physical behavior of static containers, and the thresholds for these rules were empirically determined by optimizing counting performance on our validation set. 

Our approach introduces three key rules:
\begin{itemize}
    \item \textbf{$\mathbf{H_1}$: Minimum Track Duration Filter} - This heuristic dictates that tracks shorter than 15 frames are removed. These short-lived tracks often result from unreliable detections when the container is distant from the camera, as indicated by small bounding box areas and unstable ID assignments. Thus, removing them reduces false positives and temporal noise.
    \item  \textbf{$\mathbf{H_2}$: Temporal Track Merging Based on Gaps} - This heuristic addresses track fragmentation by assuming such gaps may still belong to the same object. Two tracks of the same class are merged if the temporal gap between them is less than or equal to 20 frames. 
    \item \textbf{$\mathbf{H_3}$: Spatial Proximity Constraint for Merging} - This heuristic extends $H_2$ by enforcing a spatial proximity condition: the distance between the last detection of the first track and the first detection of the second track must be within a maximum center distance of 0.10 normalized image units.
\end{itemize}

\subsection{Geometry-Aware Overflow Segmentation}
\label{sec:proposed_methods:subsec:geometry}
\begin{figure}[t]
    \centering
    \includegraphics[width=\linewidth]{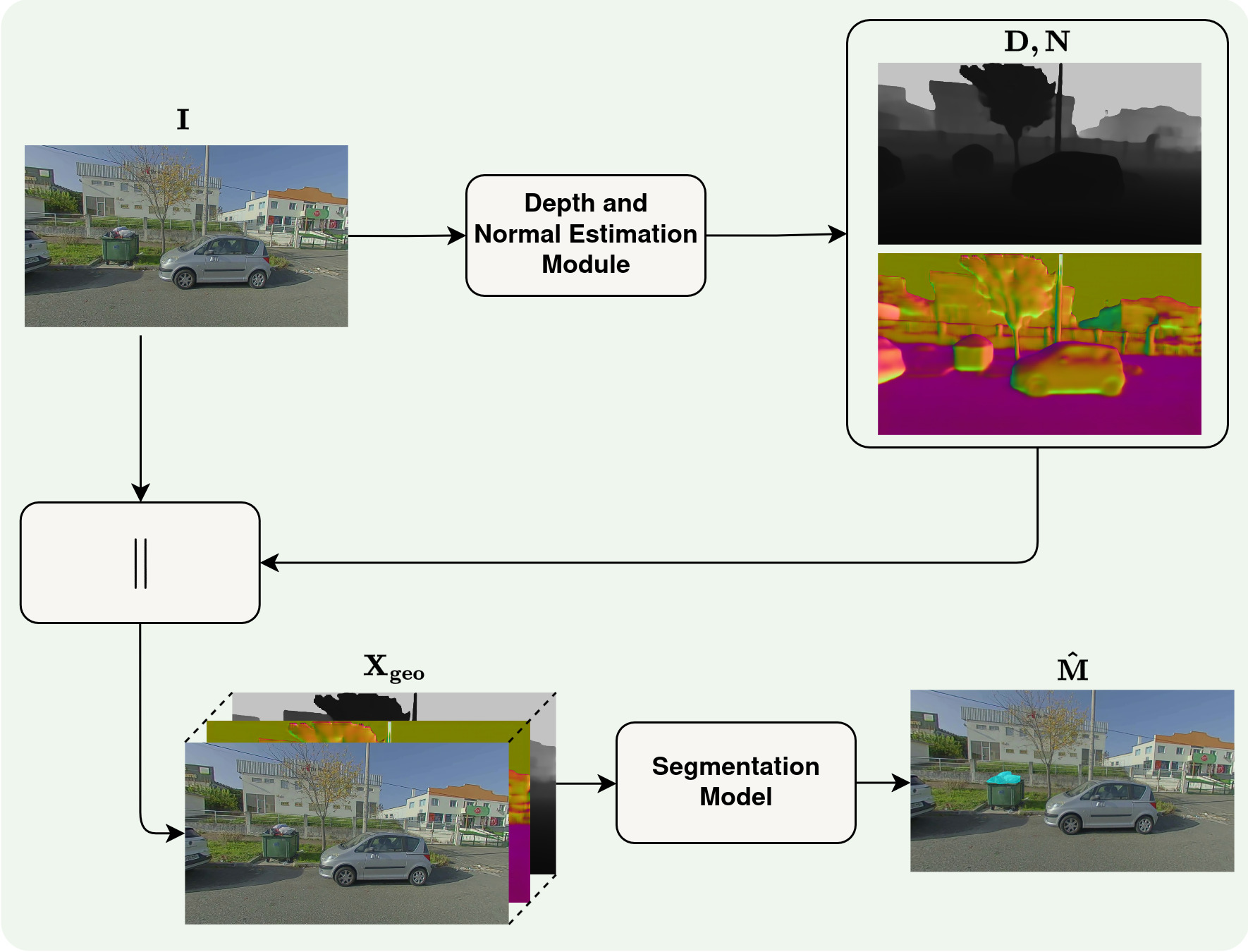}
    \caption{\textbf{Overview of the geometry-aware method for the segmentation task.} The input RGB image $I$ is processed using a geometry estimation module, which produces both a depth map $D$ and a surface normal map $N$. These are then concatenated with the original image to form an enriched input tensor $X_{\text{geo}} \in \mathbb{R}^{H \times W \times 7}$. This new representation is then fed to adapted segmentation models capable of handling multi-channel input, which output the predicted mask for overflowing waste.}
    \label{fig:proposed_method_segmentation}
\end{figure}

To compensate for the information lost during the camera's projective transform, we introduce a geometry-aware strategy to address the ambiguity of overflowing waste segmentation by leveraging 3D spatial data to supplement the 2D RGB input. As illustrated in Figure \ref{fig:proposed_method_segmentation}, our approach aims to reduce both false positives (e.g., background clutter mistaken for waste) and false negatives (missed overflow) by resolving the inherent scale ambiguity, allowing the model to differentiate between near and distant objects.

Our primary hypothesis is that depth and surface normal information can provide additional and useful cues. We use a zero-shot, single-image geometry estimation model, Metric3Dv2~\cite{hu2024metric3d45}, to infer a depth map ($D$) and a surface normal map ($N$) from each input image ($I$). Depth maps are clipped at 50~m and normalized by this value to map distances into $[0,1]$, while surface normals are stored as images and rescaled into $[0,1]$. These geometric maps are then concatenated with the original RGB channels to form an enriched 7-channel input tensor $X_{\text{geo}} = I \| N \| D \in \mathbb{R}^{H \times W \times 7}$.

\vspace{1mm}
To process this channel-extended input, the kernels of the first convolutional layer in existing segmentation architectures are adapted to accept seven input channels instead of three. The rest of the network architecture and its pretrained weights remain unchanged, allowing us to use pre-existing models with minimal modification. This fusion provides advantages as depth information helps the model distinguish foreground waste from similarly textured background elements (e.g., walls, distant vehicles) and surface normals describe the local orientation of surfaces, enabling the model to better understand the shape and form of the overflowing waste.

%% file: 5_experiments.tex
\section{Benchmarks and Results}
\label{sec:experiments}
To demonstrate the utility of StreetView-Waste and establish performance baselines, we conduct a comprehensive empirical evaluation across the three proposed tasks. This section details our experimental setup and presents a thorough analysis of the results, highlighting the unique challenges presented by our dataset. All results are reported as mean $\pm$ standard deviation computed over 10 bootstrap resamples of the test set.

\noindent\textbf{Implementation Details.}
For the three tasks, all experiments were conducted on NVIDIA GeForce RTX 5070 with 12 GB of VRAM, and the framework utilized was PyTorch~\cite{paszke2019pytorch}. Resource usage (\#Params, GFLOPs, latency/FPS at $640\times640$, and peak VRAM) is summarized in the supplementary material (Table 6).  To ensure consistency across experiments, all models were trained using a uniform image size of $640\times640$. The batch size was set to 8 for all models, except for Mask2Former~\cite{cheng2022masked32}, for which we used a reduced batch size of 4. To ensure a fair comparison, the key hyperparameters for each model (e.g., learning rate, weight decay) were independently optimized on our validation set. Regarding the dataset proportion of positive and negative samples, we employed a sampling strategy for images containing target objects versus background-only images in order to ensure a ratio of approximately 1:2 for the detection and tracking tasks and a balanced 1:1 for the overflow segmentation task. As described in~\ref{sec:proposed_methods:subsec:geometry}, depth maps were clipped at 50~m and normalized to [0,1], while surface normals were stored as images and rescaled to [0,1] before concatenation with the RGB channels.  Additionally, standard data augmentation techniques, including horizontal flipping, were applied during training to mitigate overfitting.


\subsection{Waste Container Detection}
As shown in Table \ref{tab:ap_results_detection}, the single-frame model consistently outperforms the video-based one across most categories, achieving an overall mAP@[0.5:0.95] of 0.77 versus 0.70 and an mAR@[0.5:0.95] of 0.77 versus 0.76. These results suggest that for this task, the high-quality features learned by a state-of-the-art static detector are more impactful than the temporal information exploited by DiffusionVID~\cite{roh2023diffusionvid46}. However, a key exception provides critical insight. DiffusionVID~\cite{roh2023diffusionvid46} surpasses YOLOv11~\cite{khanam2024yolov11_29} on the \textit{Battery Container} class. Our analysis suggests this is due to the unique physical properties of this class. Battery containers are typically smaller and narrower than other types, making them more susceptible to being fully occluded for several consecutive frames by street objects like poles, signs, or cars. In these scenarios, DiffusionVID's~\cite{roh2023diffusionvid46} temporal propagation mechanism can effectively fill in detections during brief occlusions where a single-frame detector would fail, demonstrating the specific conditions under which video-based models offer a distinct advantage.

\begin{table}[t]
\centering
\caption{\textbf{Benchmark of StreetView-Waste for per-class detection accuracy with the standard deviation ($\pm$ std).} The YOLOv11~\cite{khanam2024yolov11_29} model consistently outperforms DiffusionVID~\cite{roh2023diffusionvid46} across most classes, except for Battery container.}
\label{tab:ap_results_detection}
\resizebox{\linewidth}{!}{%
\begin{tabular}{@{}c ccc ccc@{}}
\toprule
\multirow{2}{*}{Class} & \multicolumn{3}{c}{DiffusionVID~\cite{roh2023diffusionvid46}} & \multicolumn{3}{c}{YOLOv11~\cite{khanam2024yolov11_29}} \\ \cmidrule(lr){2-4} \cmidrule(lr){5-7}
 & AP@0.5 & AP@[.5:.95] & AR@[.5:.95] & AP@0.5 & AP@[.5:.95] & AR@[.5:.95] \\ \midrule \midrule
Default       & 0.96$\pm$0.02 & 0.75$\pm$0.03 & 0.78$\pm$0.03 & \textbf{0.98$\pm$0.01} & \textbf{0.82$\pm$0.02} & \textbf{0.82$\pm$0.02} \\
Green         & 0.92$\pm$0.03 & 0.76$\pm$0.03 & 0.82$\pm$0.03 & \textbf{0.97$\pm$0.02} & \textbf{0.82$\pm$0.03} & \textbf{0.84$\pm$0.02} \\
Biodegradable & 0.92$\pm$0.02 & 0.75$\pm$0.02 & 0.77$\pm$0.03 & \textbf{0.96$\pm$0.01} & \textbf{0.85$\pm$0.02} & \textbf{0.86$\pm$0.02} \\
Blue          & 0.95$\pm$0.02 & 0.73$\pm$0.03 & 0.79$\pm$0.03 & \textbf{0.96$\pm$0.02} & \textbf{0.80$\pm$0.02} & \textbf{0.81$\pm$0.02} \\
Yellow        & 0.90$\pm$0.03 & 0.73$\pm$0.03 & 0.80$\pm$0.03 & \textbf{0.96$\pm$0.01} & \textbf{0.82$\pm$0.02} & \textbf{0.84$\pm$0.02} \\
Oil           & 0.92$\pm$0.02 & 0.62$\pm$0.04 & 0.69$\pm$0.04 & \textbf{0.93$\pm$0.01} & \textbf{0.70$\pm$0.03} & \textbf{0.70$\pm$0.03} \\
Battery       & \textbf{0.82$\pm$0.05} & 0.57$\pm$0.05 & \textbf{0.61$\pm$0.05} & 0.80$\pm$0.06 & \textbf{0.61$\pm$0.05} & 0.55$\pm$0.05 \\ \midrule
All             & 0.91$\pm$0.03 & 0.70$\pm$0.03 & 0.76$\pm$0.03 & \textbf{0.94$\pm$0.02} & \textbf{0.77$\pm$0.03} & \textbf{0.77$\pm$0.03} \\ \bottomrule
\end{tabular}%
}
\end{table}

\subsection{Waste Container Tracking and Counting}
This task focuses on tracking and counting waste container instances within video sequences in urban scenes, which is essential for applications such as inventory management and mapping. The results, shown in Tables \ref{tab:ap_results_tracking} and \ref{tab:ap_results_counting}, reveal significant challenges in temporal reasoning that current methods struggle with.

\begin{table}[t]
\caption{\textbf{Benchmark of StreetView-Waste for the tracking task using our post-processing heuristics.} We report the overall metrics for the baseline tracker and with the progressive application of each heuristic. The arrows ($\uparrow$/$\downarrow$) indicate whether a higher or lower value is better.}
\label{tab:ap_results_tracking}
\centering
\resizebox{\linewidth}{!}{%
\begin{tabular}{@{}ccccccc@{}}
\toprule
Model & Experiment & MOTA$\uparrow$ & IDF1$\uparrow$ & HOTA$\uparrow$ & DetA$\uparrow$ & AssA$\uparrow$ \\ \midrule \midrule
\multirow{4}{*}{ByteTrack~\cite{zhang2022bytetrack47}}
& Baseline                        & 76.80\% & 81.40\% & 69.76\% & 69.96\% & 69.83\% \\
& $H_1$ (Duration)                & 77.10\% & \textbf{82.20\%} & 69.73\% & 69.30\% & \underline{70.40\%} \\
& $H_1$ + $H_2$ (Temporal)        & 77.00\% & 55.10\% & 50.98\% & 45.62\% & 57.03\% \\
& $H_1$ + $H_2$ + $H_3$ (Spatial) & 77.20\% & 66.80\% & 59.60\% & 56.35\% & 63.15\% \\ \midrule \midrule
\multirow{4}{*}{BoT-SORT~\cite{botsort}}
& Baseline                        & \textbf{82.50\%} & 79.60\% & \underline{71.50\%} & \textbf{75.27\%} & 68.17\% \\
& $H_1$ (Duration)                & \underline{82.40\%} & \underline{82.10\%} & \textbf{72.09\%} & \underline{73.80\%} & \textbf{70.61\%} \\
& $H_1$ + $H_2$ (Temporal)        & \textbf{82.50\%} & 75.80\% & 67.34\% & 65.82\% & 69.06\% \\
& $H_1$ + $H_2$ + $H_3$ (Spatial) & \textbf{82.50\%} & 79.80\% & 70.30\% & 70.69\% & 70.06\% \\ \bottomrule
\end{tabular}%
}
\end{table}

\begin{figure*}
    \centering
    \includegraphics[width=.95\linewidth]{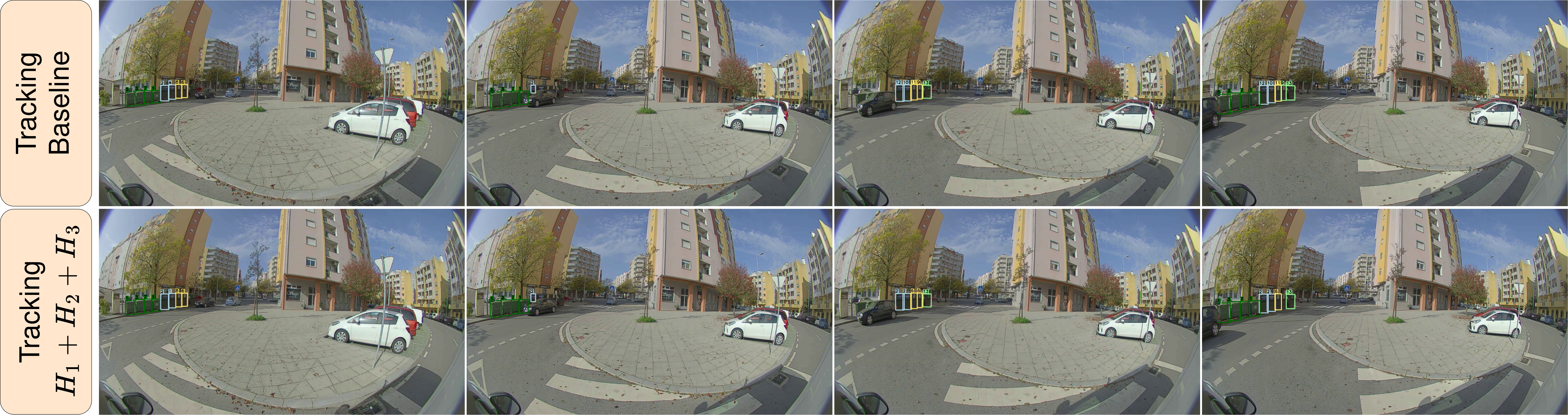}
    \caption{\textbf{Qualitative results for the multi-object tracking.} This scenario, common in our dataset, shows the difficulty of keeping track and explains the results of the lower IDF1 score when introducing temporal heuristics. This improves track continuity but corrupts identity (lowering IDF1).}
    \label{fig:visual_examples_track}
\end{figure*}

The comparison between ByteTrack~\cite{zhang2022bytetrack47} and BoT-SORT~\cite{botsort} highlights the dataset's difficulty. BoT-SORT achieves a stronger baseline with 82.5\% of MOTA and 71.5\% of HOTA, which reflects its improved detection association. However, both trackers suffer from frequent identity switches and degraded performance once heuristics are introduced. For example, while our spatio-temporal heuristics ($H_1+H_2+H_3$) reduce ByteTrack’s counting error drastically (MAE from 3.48 to 0.71, with SAD dropping from 73 to 15), they simultaneously reduce identity preservation (IDF1 81.4\% $\rightarrow$ 66.8\%, HOTA 69.8\% $\rightarrow$ 59.6\%). A similar pattern occurs with BoT-SORT, where heuristics improve counting accuracy (MAE 6.43 $\rightarrow$ 0.90) but disrupt balanced association quality. A detailed breakdown of these counting improvements for each class is available in the supplementary material (Tables 4 and 5). As illustrated in Figure \ref{fig:visual_examples_track}, this happens frequently when containers of the same type appear sequentially along a collection route. Therefore, we interpret these heuristics not as improvements but as diagnostic tools; they show that even simple, rule-based temporal or spatial constraints expose the fundamental difficulty of maintaining consistent identities in cluttered street-level scenes. This confirms that container tracking under realistic operational conditions remains an open challenge, and StreetView-Waste provides a benchmark for studying these limitations in depth.

\begin{table}[t]
\centering
\caption{\textbf{Benchmark of StreetView-Waste for object counting accuracy.} The results evidence strong performance when using our proposed heuristics.}
\label{tab:ap_results_counting}
\resizebox{\linewidth}{!}{%
\begin{tabular}{@{}cccccc@{}}
\toprule
Model & Experiment & MAE$\downarrow$ & SAD$\downarrow$ & RMSE$\downarrow$ & MAPE$\downarrow$ \\ \midrule \midrule
\multirow{4}{*}{ByteTrack~\cite{zhang2022bytetrack47}}
& Baseline     & 3.48 & 73  & 7.80 & 82.96\% \\
& $H_1$ (Duration) & 1.05 & 22  & 1.93 & 24.96\% \\
& $H_1$ + $H_2$ (Temporal) & \underline{0.76} & \underline{16} & 1.75 & \underline{17.77}\% \\
& $H_1$ + $H_2$ + $H_3$ (Spatial) & \textbf{0.71} & \textbf{15} & \underline{1.70} & \textbf{16.03\%} \\ \midrule \midrule
\multirow{4}{*}{BoT-SORT~\cite{botsort}}
& Baseline     & 6.43 & 135 & 9.60 & 187.61\% \\
& $H_1$ (Duration) & 1.19 & 25 & 1.91 & 27.19\% \\
& $H_1$ + $H_2$ (Temporal) & 0.81 & 17 & \textbf{1.36} & \textbf{16.03}\% \\
& $H_1$ + $H_2$ + $H_3$ (Spatial) & 0.90 & 19 & 1.79 & 17.96\% \\ \bottomrule
\end{tabular}%
}
\end{table}

\begin{figure*}[ht]
    \centering
    \includegraphics[width=.95\linewidth]{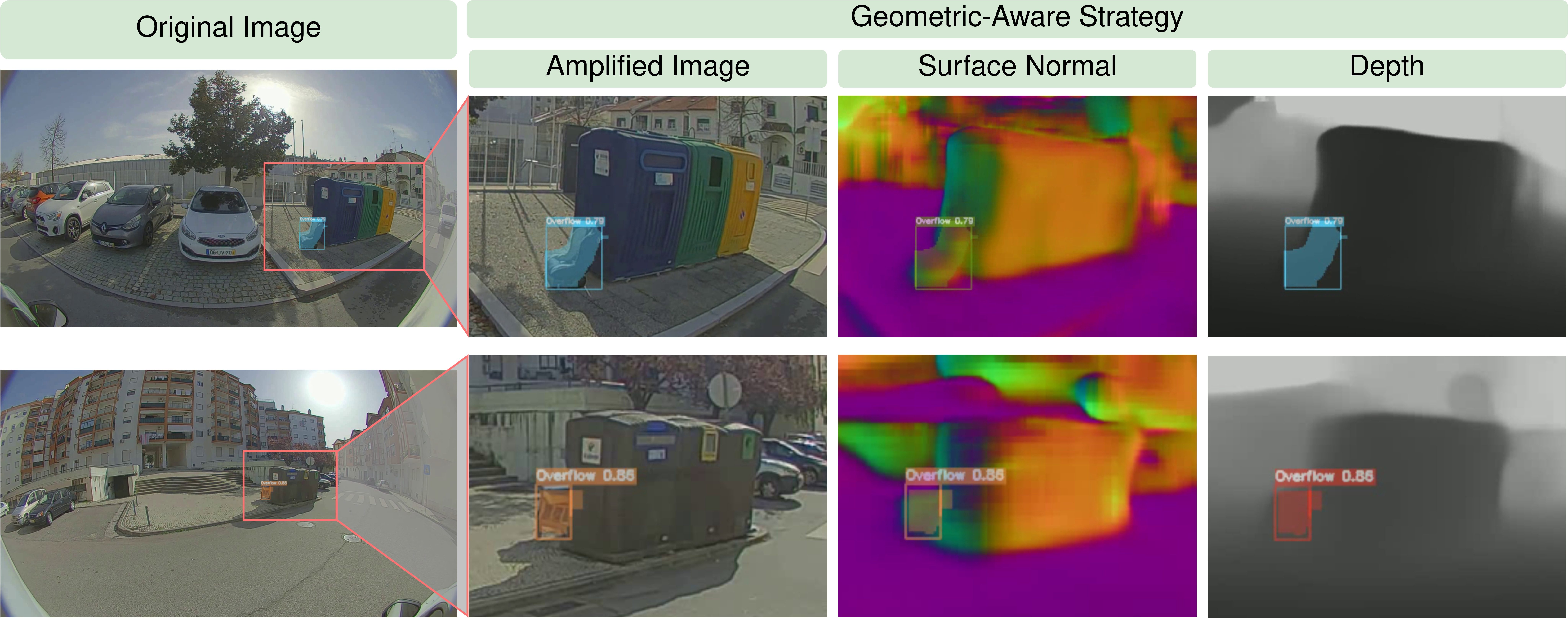}
    \caption{\textbf{Qualitative results for the waste overflow segmentation task, comparing the original images with our proposed geometry-aware strategy.} The columns show, from left to right: the original input image, the same image with the segmentation result from our method, the estimated surface normal map, and the estimated depth map.}
    \label{fig:visual_examples_seg}
\end{figure*}

\subsection{Waste Overflow Segmentation}
For this task, we evaluated five state-of-the-art instance segmentation models to establish baselines for the challenging task of segmenting amorphous, overflowing waste. Table \ref{tab:ap_results_all} compares the performance of each model on the standard RGB fisheye projection~\cite{kannala2006generic} against our proposed geometry-aware strategy, which incorporates geometric priors (depth and surface normals).
\begin{table}[t]
\caption{\textbf{Benchmark of StreetView-Waste for the waste overflow segmentation task using our geometry-aware strategy.} The addition of geometric priors improves performance for the majority of the models.}
\label{tab:ap_results_all}
\centering
\resizebox{\linewidth}{!}{%
\begin{tabular}{@{}c@{\hspace{1em}}c c c c c@{}}
\toprule
& Model & Experiment & mAP@0.5 & mAP@[0.5:0.95] & B-IoU \\ 
\midrule \midrule

\multirow{6}{*}{%
    \rotatebox{90}{%
        \begin{tabular}{@{}c@{}}\textbf{Complex} \\ \textbf{Architectures}\end{tabular}%
    }%
} 
& \multirow{2}{*}{SOLOv2~\cite{wang2020solov2_31}} & Baseline (RGB) & \textbf{0.20$\pm$0.03} & \textbf{0.10$\pm$0.01} & \textbf{0.30$\pm$0.02} \\
& & Ours (Geometric Cues) & 0.07$\pm$0.01 & 0.03$\pm$0.00 & 0.20$\pm$0.01 \\ 
\cmidrule(l){2-6}
& \multirow{2}{*}{Mask2Former~\cite{cheng2022masked32}} & Baseline (RGB Data) & 0.18$\pm$0.02 & 0.11$\pm$0.01 & \textbf{0.38$\pm$0.02} \\
& & Ours (Geometric Cues) & \textbf{0.29$\pm$0.02} & \textbf{0.13$\pm$0.01} & 0.32$\pm$0.02 \\ 
\cmidrule(l){2-6}
& \multirow{2}{*}{Mask R-CNN~\cite{he2017mask27}} & Baseline (RGB) & \textbf{0.41$\pm$0.02} & \textbf{0.26$\pm$0.01} & \textbf{0.54$\pm$0.02} \\
& & Ours (Geometric Cues) & 0.12$\pm$0.01 & 0.06$\pm$0.01 & 0.38$\pm$0.01 \\ 
\midrule \midrule

\multirow{4}{*}{%
    \rotatebox{90}{%
        \begin{tabular}{@{}c@{}}\textbf{Lightweight} \\ \textbf{Models}\end{tabular}%
    }%
} 
& \multirow{2}{*}{YOLACT~\cite{bolya2019yolact28}} & Baseline (RGB) & 0.41$\pm$0.02 & 0.22$\pm$0.01 & 0.87$\pm$0.02 \\
& & Ours (Geometric Cues) & \textbf{0.52$\pm$0.02} & \textbf{0.31$\pm$0.02} & \textbf{0.90$\pm$0.01} \\ 
\cmidrule(l){2-6}
& \multirow{2}{*}{YOLOv11~\cite{khanam2024yolov11_29}} & Baseline (RGB) & 0.50$\pm$0.01 & 0.30$\pm$0.01 & \textbf{0.77$\pm$0.01} \\
& & Ours (Geometric Cues) & \textbf{0.52$\pm$0.01} & \textbf{0.31$\pm$0.01} & \textbf{0.77$\pm$0.01} \\

\bottomrule
\end{tabular}%
}
\end{table}

The results reveal an architectural division in the ability to use multi-modal information for this task. On the one hand, lightweight models show significant performance gains with the added geometric data, with YOLACT~\cite{bolya2019yolact28} achieving the highest overall mAP@[0.5:0.95] and B-IoU of 0.31 and 0.90, respectively. On the other hand, more complex models exhibit performance degradation. Our analysis suggests that this divergence occurs due to fundamental architectural limitations when faced with unstructured targets like litter. The models that successfully adapt, such as YOLACT~\cite{bolya2019yolact28} and Mask2Former~\cite{cheng2022masked32}, possess more flexible instance representation mechanisms, such as YOLACT's~\cite{bolya2019yolact28} prototype-based design and Mask2Former's~\cite{cheng2022masked32} transformer-based cross-attention mechanism, which allows them to effectively learn relationships between modalities across the entire scene.
In contrast, models like Mask R-CNN~\cite{he2017mask27} are built on rigid structural assumptions that are violated by our dataset's challenges. Two-stage models rely on a Region Proposal Network (RPN) that fails when there are no well-defined, "object-like" structures to propose from the amorphous litter. Similarly, SOLOv2's~\cite{wang2020solov2_31} grid-based design, which assumes one compact instance per grid cell, cannot handle waste that expands across multiple grid cells. For these architectures, the introduction of new geometric channels does not provide a helpful signal but instead exposes and amplifies these core weaknesses, leading to a significant degradation in performance. Furthermore, Figure~\ref{fig:visual_examples_seg} depicts original images processed using the geometry-aware method, and to complement, \textbf{ablation studies} are provided in the supplementary material (Table 3), analyzing the individual contributions of depth and surface normal cues.

\subsection{Discussion and Limitations}
Our empirical evaluation establishes baselines for StreetView-Waste and highlights key challenges for future research. While geometric priors improve segmentation, the approach remains sensitive to errors in the predicted depth and surface normals, which can degrade performance in cluttered or low-contrast scenes. Moreover, complex architectures such as Mask R-CNN~\cite{he2017mask27} and SOLOv2~\cite{wang2020solov2_31} did not benefit from the added modalities, suggesting that larger training sets or more specialized multimodal fusion blocks (e.g., cross-attention mechanisms) may be required. These observations point to a need for further exploration of how geometric cues interact with different architectural designs.

%% file: 6_conclusions.tex

\section{Conclusion}
\vspace{-2mm}
We introduced StreetView-Waste, a large-scale dataset for benchmarking detection, tracking, and overflow segmentation in urban waste management. We applied two simple strategies, not as novel solutions but as diagnostic tools, to probe the limitations of state-of-the-art models. Our experiments yield two key insights: (1) simple heuristics markedly reduce counting errors, revealing that current trackers lack fundamental temporal reasoning; and (2) our geometric fusion strategy lowers the performance of complex architectures, showing their fragility to multi-modal inputs and amorphous objects. These results confirm StreetView-Waste as a critical diagnostic benchmark. Future work will extend the dataset with GPS to enable logistics-oriented applications.
\\ \\
\section{Acknowledgments}
\vspace{-2mm}
This work is funded by national funds through FCT – Fundação para a Ciência e a Tecnologia, I.P., and, when eligible, co-funded by EU funds under project/support UID/50008/2025 – Instituto de Telecomunicações, with DOI identifier \url{https://doi.org/10.54499/UID/50008/2025}
This work is also financed by the project WATERMARK\footnote{WATERMARK project (Watermark-Based Algorithms for Trustworthy Media Authentication and Robust Certification in Public Administration), Project No. 2024.07356.IACDC, supported by “RE-C05-i08.M04 – Support the launch of a program of R\&D projects aimed at the development and implementation of advanced systems in cybersecurity, artificial intelligence, and data science in public administration, as well as a scientific training program,” under the Recovery and Resilience Plan (PRR), as part of the funding agreement signed between the Recovery Portugal Task Force (EMRP) and the Foundation for Science and Technology (FCT).} and supported by UID/04516/NOVA Laboratory for Computer Science and Informatics (NOVA LINCS) with the financial support of FCT.IP.